\documentclass[10pt,twocolumn,letterpaper]{article}

\usepackage{wacv}              

\usepackage{multirow}
\usepackage{float}
\usepackage{longtable}
\usepackage{makecell} 

\usepackage{colortbl}  
\usepackage{xcolor}    
\usepackage{soul}

\definecolor{wacvblue}{rgb}{0.21,0.49,0.74}
\usepackage[pagebackref,breaklinks,colorlinks,allcolors=wacvblue]{hyperref}

\def\wacvPaperID{186} 
\def\confName{WACV}
\def\confYear{2026}

\title{FCC: Fully Connected Correlation for One-Shot Segmentation}

\author{Seonghyeon Moon\\
Roblox, Brookhaven National Laboratory\\
{\tt\small smoon@roblox.com}
\and
Haein Kong\\
Rutgers University\\
{\tt\small 
haein.kong@rutgers.edu}
\and
Muhammad Haris Khan\\
Mohamed Bin Zayed University of Artificial Intelligence\\
{\tt\small 
muhammad.haris@mbzuai.ac.ae}
\and
Mubbasir Kapadia\\
Roblox\\
{\tt\small 
mkapadia@roblox.com}
\and
Yuewei Lin\\
Brookhaven National Laboratory\\
{\tt\small ywlin@bnl.gov}
}

\begin{document}
\maketitle
\begin{abstract}

One-shot segmentation~(OSS) aims to segment the target object in a query image using only one set of support image and mask. Therefore, having strong prior information for the target object using the support set is essential to guide the initial training of OSS, which leads to the success of one-shot segmentation in challenging cases, such as when the target object shows considerable variation in appearance, texture, or scale across the support and query images. To enrich this prior knowledge, we introduce FCC~(Fully Connected Correlation) which integrates pixel-level correlations between support and query features, capturing associations that reveal target-specific patterns and correspondences in both same-layers and cross-layers. FCC captures previously inaccessible target information, effectively addressing the limitations of support mask. Our approach consistently demonstrates state-of-the-art performance in the PASCAL, COCO, and domain shift tests, while also notably accelerating model convergence. We conducted an ablation study and cross-layer correlation analysis to validate FCC's core methodology. These findings reveal the effectiveness of FCC in enhancing prior information and overall model performance for OSS\footnote{Our code is available at: https://github.com/moonsh/FCC-Fully-Connected-Correlation-for-Few-Shot-Segmentation}.


\end{abstract}    
\section{Introduction}
\label{sec:intro}

\begin{figure}[!ht]
  \centering
    \includegraphics[height=6.5cm]{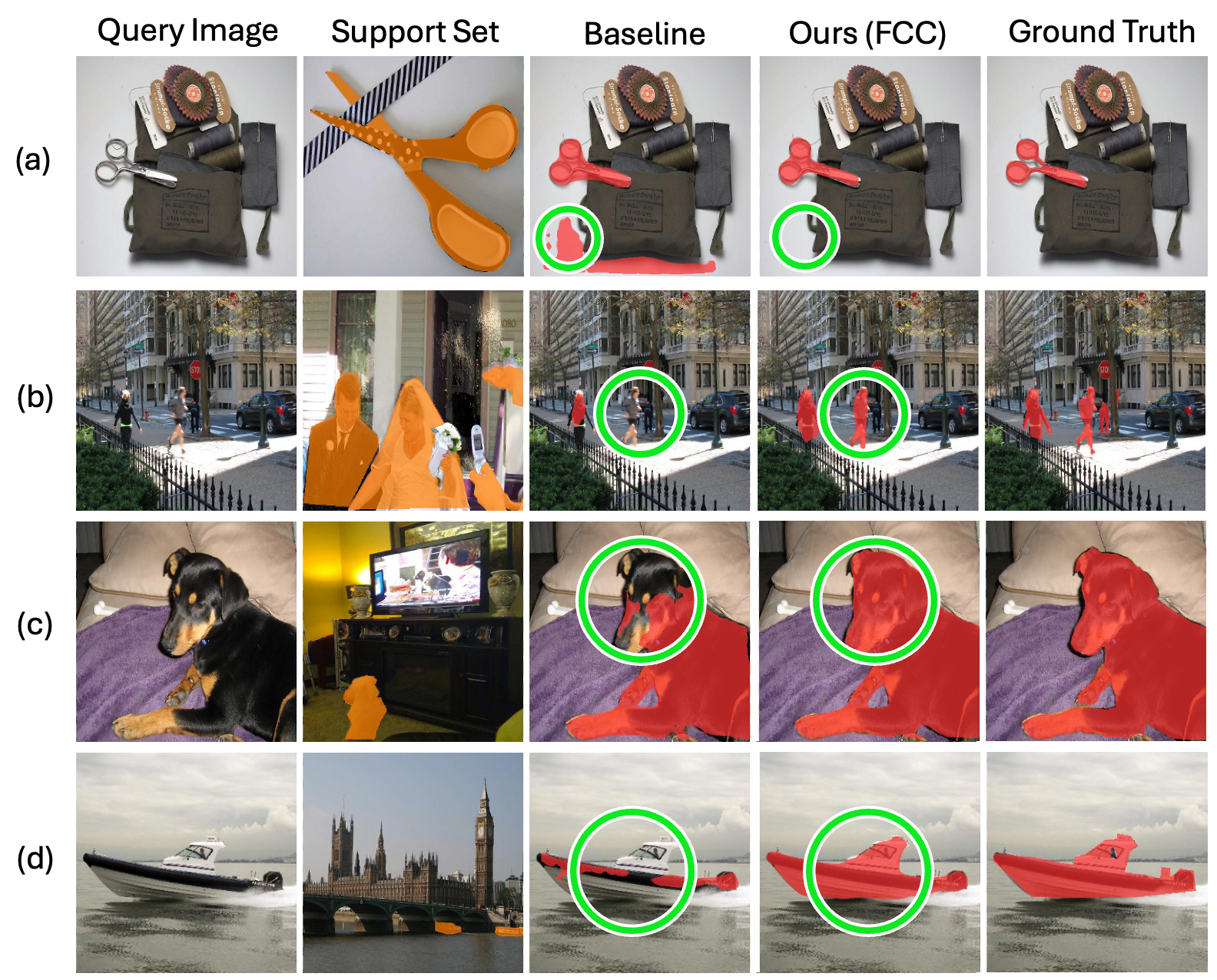}

   \caption{The proposed method, FCC, demonstrates precise segmentation in challenging scenarios on COCO-20$^i$~\cite{lin2015microsoft} compared to baseline: (a) \textcolor{orange}{Scale Difference}: where the target object appears at different scales in the query and support images; (b) \textcolor{orange}{Scale Difference} + \textcolor{purple}{Occlusion}: where the target object is partially occluded by other objects, with only part of it visible in the support set; (c) \textcolor{orange}{Scale Difference} + \textcolor{purple}{Occlusion} + \textcolor{teal}{Shape Difference}: where objects of the same class differ in color and shape, as seen in various dog breeds; (d) \textcolor{orange}{Scale Difference} + \textcolor{purple}{Occlusion} + \textcolor{teal}{Shape Difference} + \textcolor{red}{Limited Information}: the most challenging case, where minimal information is provided, and all previous challenges are present simultaneously. The green circle highlights the area of significant improvement with FCC.} 
   
   
   \label{fig:4cases}
   \vspace{-1.5em} 

\end{figure}
The advancement of deep learning has significantly improved object segmentation~\cite{FCN, Deeplab, pyramidscene}, enabling its application across diverse fields such as healthcare, biology, semiconductor manufacturing, and autonomous driving. The improvements in object segmentation have largely been achieved through supervised learning. Despite these advances, fully supervised segmentation methods face limitations in many of these fields due to data scarcity and the challenges of acquiring high-quality, annotated datasets across varied conditions and specialized applications~\cite{song2023comprehensive, alzubaidi2023scarcity}. 
Few-shot learning has emerged as an active area of research to address this problem, aiming to achieve strong performance with a small amount of data~\cite{snell2017prototypical,wang2020generalizingexamplessurveyfewshot, song2023comprehensive}. By leveraging few-shot learning, the substantial costs associated with data acquisition and labeling can be reduced, significantly expanding the potential for segmentation applications in fields like biology, electronics, and satellite imaging~\cite{lin2023few_micai,fewshot_elec,fewshot_satelite}.

In One-shot segmentation (OSS), only one example with target object mask is provided. Using this limited information, we need to segment the target object within a new image, known as the query image. This task is believed to be introduced by~\citet{OSLSM} for the first time. Since then, numerous approaches have been proposed. Most previous OSS works utilize pixel-level correlation between query and support features to generate prior mask or aggregate support information \cite{rethinking, HSNet, ASNet, vrpsam, dscm, hmmasking,msi, VAT}. 


Most previous works generate a prior mask based on a correlation map, which serves as a key component by providing strong prior information to the network. Pixel-level correlation in previous methods is mostly computed comparing features from the same layer in a network~\cite{msi,hmmasking,VAT,HSNet,ASNet,piclip,vrpsam}. However, relying only on same-layer features could miss nuanced details about target information, especially when objects' appearance, scale, or structure in the support and query images differ considerably. This could lead to a performance bottleneck in OSS~(Fig.~\ref{fig:4cases}). 

\begin{figure}
 \centering 
 \includegraphics[width=22em]{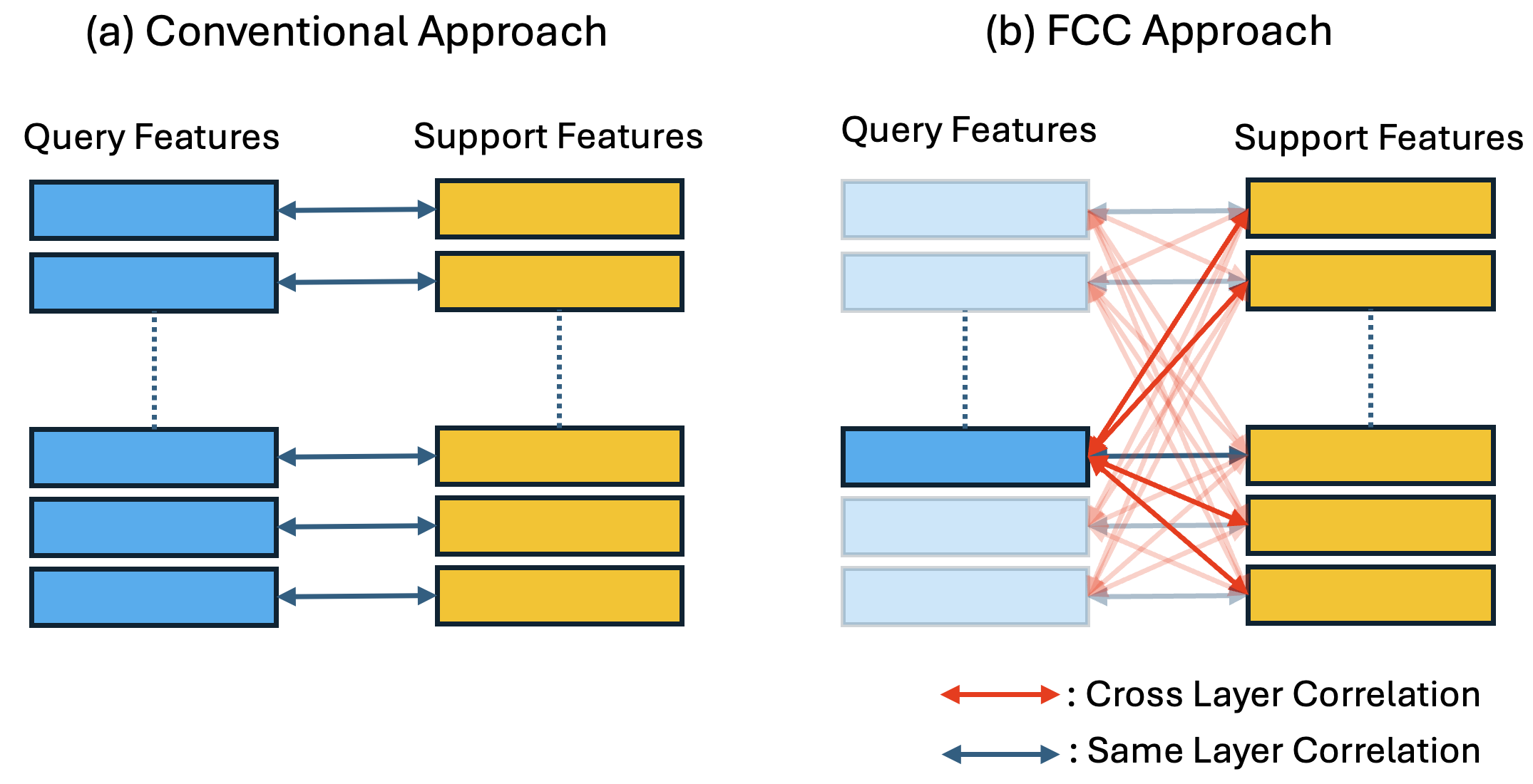}
 \caption{(a) Previous methods in few-shot segmentation typically compare features at the same-level layer to generate the correlation map. (b) FCC, however, leverages all layers and their correlations to capture maximum target information. Note: For clarity, only one query feature block is visualized.}
 \label{fig:comparison_conv_fcc}

\end{figure}

Despite the potential benefits that cross-layer can bring, cross-layer correlations for generating pixel-level correlations have been underexplored in previous methods~(Fig.~\ref{fig:comparison_conv_fcc}). 
%
Comparing features across different layers has often been considered counterintuitive. In convolutional neural networks~(CNN) such as ResNet~\cite{Resnet}, lower layers are understood to capture low-level image features, while higher layers progressively capture more abstract information. This means each layer represents distinct types of information, making comparing same-layer features reasonable. Therefore, this same-layer pixel-level correlation is considered as \textit{defacto} in OSS. This concept is adopted to the Vision Transformer~(ViT)~\cite{vit} architecture, with the assumption that ViT exhibits a similar hierarchical understanding.  
%
%
%
%


However, \citet{raghu2021vision} observed significant differences between ViT~\cite{vit} and CNN. ViT captures more uniform representations with greater similarity between lower and upper layers than CNN. Therefore, features from ViT are less hierarchical than convolutional neural networks, making comparisons across different ViT layers a reasonable approach for uncovering detailed information at various levels. Also, the architecture of ViT allows to compute cross-layer correlation directly. Features from each layer in ViT maintain the same width and height with the same channels, making comparisons between different layers possible. 



In this paper, we present Fully Connected Correlation~(FCC), a novel approach designed to enhance one-shot segmentation by leveraging cross-layer correlations in addition to traditional same-layer comparisons. Unlike prior methods that typically rely on correlation maps generated from features at the same layer, FCC integrates correlations from multiple levels of the Vision Transformer~(ViT) encoder. This enables FCC to capture a richer and more nuanced understanding of the target object, improving segmentation in complex scenarios where the appearance, scale, or structure of the object in the query image differs significantly from the support images. We summarize our key contributions as follows:




\begin{itemize}[topsep=-3pt, noitemsep]

    \item  We propose fully connected correlation~(FCC). FCC captures rich cross-layer information, enabling a network to integrate target features across multiple layers. This approach enhances the model's ability to construct a more comprehensive and robust feature representation.
   
    \item We design a Dual-Conditioned Fully Connected Correlation~(DCFC) to enhance the effectiveness of FCC. DCFC enables the detection of hidden target information not present in the support mask, and FCC enhances this capability even further.
    
    \item Extensive experiments and analyses validate the effectiveness of FCC. In PASCAL-5$^i$~\cite{pascal}, COCO-20$^i$~\cite{lin2015microsoft}, and domain shift test, FCC performs significantly better than the existing state-of-the-art.

    \item FCC establishes a strong prior information. As a result, FCC accelerates model convergence by up to 2.7 times on PASCAL-5$^i$ and COCO-20$^i$.
    
\end{itemize}

\section{Related Works}
\label{sec:Related}
\begin{figure*}
  \centering
    \includegraphics[height=6.2cm]{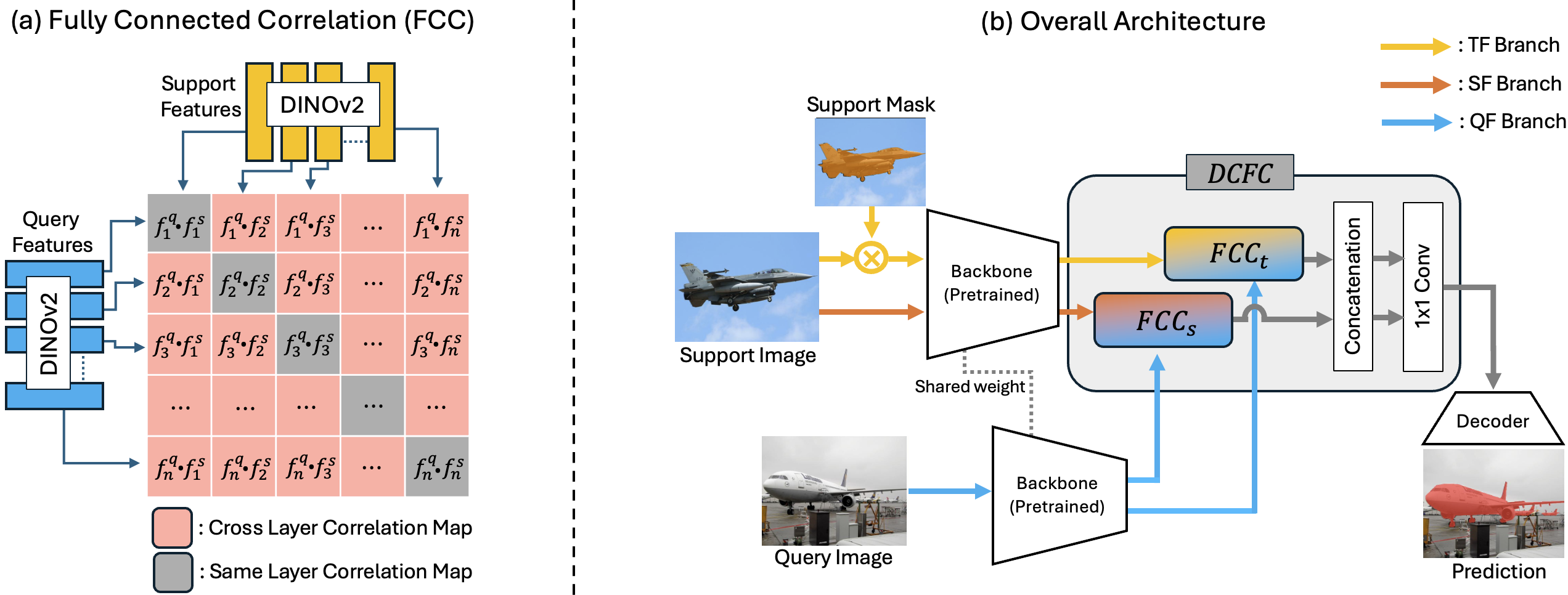}
    \caption{(a)~The Fully Connected Correlation~(FCC) integrates cross-layer and same-layer correlations to capture comprehensive information. (b)~The support image and the masked support image are processed through the backbone to extract support and target features, which are then used to generate fully connected correlation maps, $FCC_t$ and $FCC_s$. Query features extracted from the query image contribute to calculating these correlation maps, which are then concatenated and passed through a 1x1 convolution layer to reduce the channel size. The resulting output is fed into the decoder to predict the target segmentation.} 
    
    \label{fig:method}
\end{figure*}

\noindent\textbf{One-Shot Segmentation (OSS):} 
One-shot segmentation aims to perform pixel-level classification of target objects using only one annotated support example. OSLSM~\cite{OSLSM} is considered the first method aiming to solve the OSS problem. OSLSM proposed a model consisting of a condition branch and a segmentation branch. Many follow-up studies have proposed various methods to address OSS~\cite{Co-FCN,AMP-2,PMM,FWB,PFENet,PANet,CANet,ASGNet,DAN,FSOT,Zhang2020SGOneSG,CWT,HSNet,ASNet,BAM,VAT2,hmmasking}.

\noindent \textbf{Vision Foundation Models:} 
In recent years, vision foundation models have emerged in the field of computer vision and demonstrated impressive performance on various vision tasks. For example, SAM~\cite{SAM} enables interactive image segmentation based on users' prompts. As SAM showed robust zero-shot performance on segmentation tasks, researchers leveraged it to improve OSS performance. Matcher~\cite{liu2023matcher} leveraged SAM to generate mask proposals for selecting high-quality masks, while VRP-SAM~\cite{vrpsam} focused on training a network to generate prompts by learning optimal prompts for segmentation tasks. 
Painter~\cite{Painter} leverages in-context learning by using redefinition of the output space of tasks as images and masked image modeling framework. SegGPT~\cite{seggpt} is another generalist model for image segmentation based on Painter. In addition, Contrastive Language-Image Pretraining (CLIP)~\cite{clip} is a multi-modal foundation model that jointly uses an image and text encoder with contrastive learning. CLIP showed strong zero-shot performance on a variety of image classification tasks, showing potential uses in downstream vision tasks like FSS. For instance, PI-CLIP~\cite{piclip} generates reliable prior information like visual-text and visual-visual prior information based on CLIP. LLaFS~\cite{llafs} leverages a text and image encoder of CLIP to employ large language models for OSS tasks. Recently, DSCM-DFN~\cite{dscm} extracts implicit knowledge from vision and vision-language foundation models to obtain fine-grained segmentation, finding the combination of DINOv2 and DFN is especially beneficial for OSS. In contrast, our method integrates frozen pre-trained vision transformers without relying on SAM or CLIP, enabling FCC to generalize more flexibly to unseen classes.

\noindent\textbf{Correlation in OSS:} 
Pixel-level correlation between query features and support features can extract segmentation cues and generate prior masks. Prior
Mask Guidance (PMG) is a method first proposed by \citet{tian2020prior}, using pixel-level correlation to generate a prior mask. PMG has been widely adopted in subsequent OSS methods~\cite{HSNet,VAT,hmmasking,msi,rethinking,vrpsam,piclip,dscm,ASNet}, leaving correlation as a powerful and important tool in few-shot segmentation tasks. HSNet~\cite{HSNet} introduced a method called hypercorrelation, which leverages a rich set of features from the intermediate layers. These complex correlation maps are then processed using a 4D convolutional model. Subsequently, VAT~\cite{VAT} further utilizes these correlation maps through a 4D convolutional transformer. HM~\cite{hmmasking} and MSI~\cite{msi} also leveraged hypercorrelation approach in OSS tasks. Recently, ~\citet{rethinking} proposed adaptive buoys correlation to mitigate false matches by using reference features (buoys) and buoys-level correlation for OSS.
Apart from these methods, correlation maps serve as initial prior information in training networks in many OSS approaches~\cite{vrpsam,piclip}. 

However, we observe that the previous models only used the same-layer correlation between query and support features. This practice could be suboptimal for Vision transformer architecture as it could lose the prior knowledge existing in cross-layer correlation, especially challenging scenarios~(Fig.~\ref{fig:4cases}). To address this issue, we propose Fully Connected Correlation (FCC) to obtain a comprehensive and informative prior mask to improve OSS performance.



\section{Overall Method}


\begin{figure*}[!htp]
  \centering
    \includegraphics[height=9.1cm]{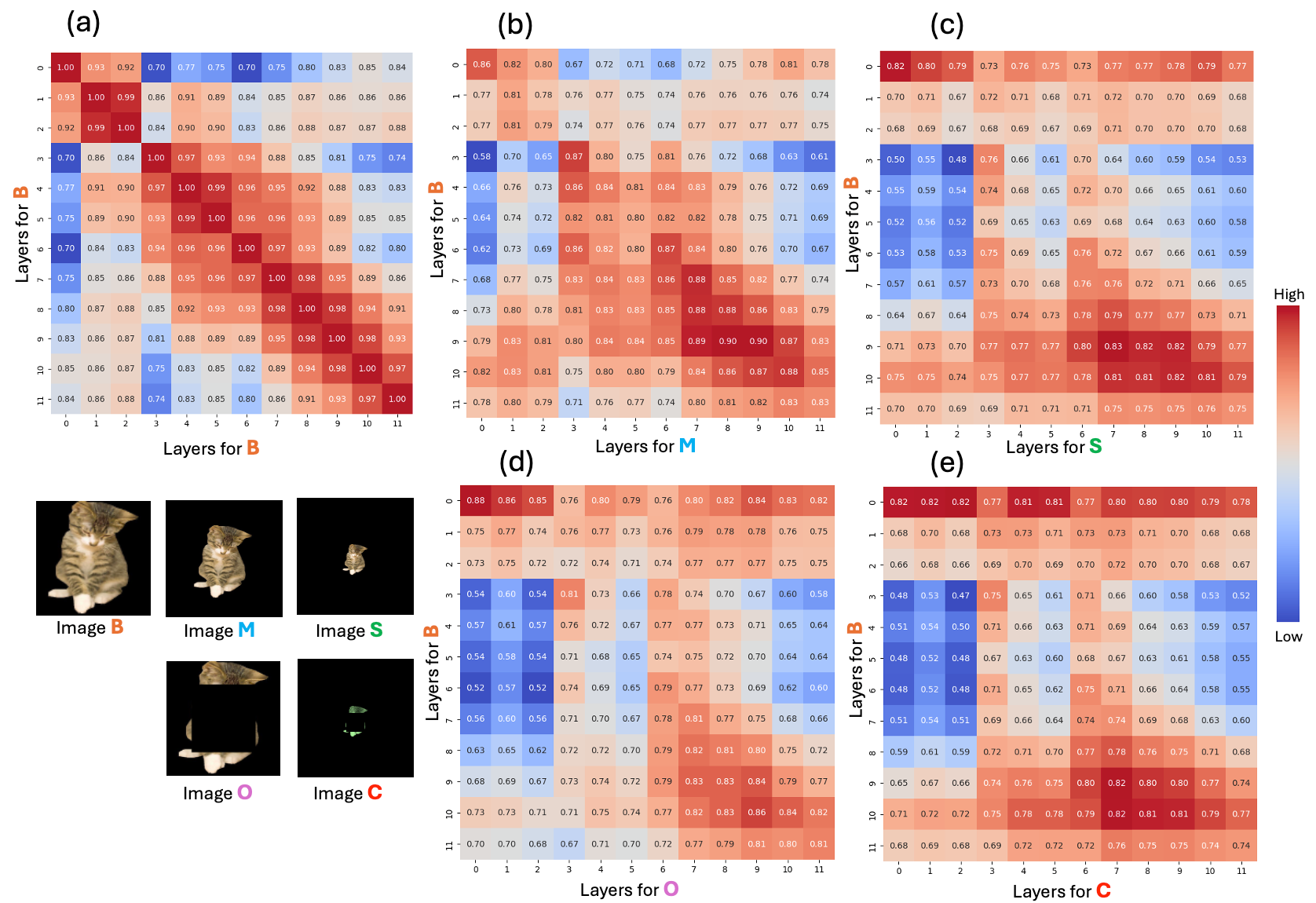}
    \caption{The Centered Kernel Alignment~(CKA~\cite{pmlr-v97-kornblith19a,cka2}) similarity heatmaps between features from DINOv2-B layers illustrate the effect of target object differences. The background of the images is removed to focus solely on the target object. (a)~The CKA displays a stronger concentration along the diagonal in the entire heatmap because we compare the same features. (b)~shows the CKA between the biggest and middle-size cat. The center to the right bottom diagonal area is highlighted. (c)~shows the similarity between the biggest and the smallest cat. High-layer areas are concentrated along the diagonal, while low-to-mid layers appear to spread. (d) We simulate an occlusion scenario where the cat is obstructed by a large square. (e) We replicate the most challenging conditions, including scale variation, occlusion, shape differences, and limited information. \textbf{When extracting features, low to middle layers capture different levels of detail based on variations in the object, while higher layers tend to capture more abstract, semantic representations.} This results in high similarity values appearing outside of the diagonal entries. }
    
    \label{fig:cka_comp_scalediff}
\end{figure*}

Fig.~\ref{fig:method} illustrates the overall architecture of our proposed method (FCC). In this architecture, DINOv2~\cite{oquab2023dinov2} serves as the backbone for feature extraction. We extract features from two images: a masked support image and a support image. These features are used to calculate correlation maps with query features. Our method integrates both same-layer and cross-layer correlation maps between support and query features to capture a comprehensive and detailed understanding of the target object's structure, scale variations, and contextual relationships. All correlation maps are concatenated and passed through a 1x1 convolution layer to reduce channel size.

\subsection{Problem Setting}

We aim to build a model that can segment the target object in a query image using only one annotated example from the target class.
Following previous approaches in OSS~\cite{HSNet, ASNet, VAT2}, we adopt an episodic training strategy for our model. It is assumed that we are provided with two distinct image sets, $\mathcal{C}_{\text{train}}$ and $\mathcal{C}_{\text{test}}$, representing the training and testing classes, respectively. 
The training dataset $\mathcal{D}_{\text{train}}$ is drawn from $\mathcal{C}_{\text{train}}$, while the testing dataset $\mathcal{D}_{\text{test}}$ is sampled from $\mathcal{C}_\text{{test}}$.  Multiple episodes are constructed from both $\mathcal{D}_\text{{train}}$ and $\mathcal{D}_\text{{test}}$. Each episode consists of a support set, $\mathcal{S} = (I^s, M^s)$, and a query set, $\mathcal{Q} = (I^q, M^q)$, having a pair of an image and mask where $I^s$ refers to an image, and $M^s$ is its corresponding mask. 
Consequently, $\mathcal{D}_\text{{train}} = \left\{(\mathcal{S}_i, \mathcal{Q}_i) \right\}_{i=1}^{N_\text{{train}}}$ and $\mathcal{D}_\text{{test}} = \left\{(\mathcal{S}_i, \mathcal{Q}_i) \right\}_{i=1}^{N_\text{{test}}}$, where $N_\text{{train}}$ and $N_\text{{test}}$ represent the number of episodes, used for training and testing. 
During training, we repeatedly sample episodes from $\mathcal{D}_\text{{train}}$ which allows the model to learn a mapping from $(I^s, M^s, I^q)$ to the query mask $M^q$. After training the model, we evaluate its one-shot segmentation performance on $\mathcal{D}_\text{{test}}$ by randomly sampling episodes without further optimization. 


\subsection{Feature Extraction}
We employed a transformer-based backbone for our method. For clarity, we selected the DINOv2~\cite{oquab2023dinov2} backbone to explain our method. 
ViT-B/14 model contains 12 transformer layers. The entire 12 layers are used to extract image features. Let \( L = \{ l_1, l_2, \ldots, l_{12} \} \) represent each of the ViT-B layers. Intermediate features \( f_l \) were extracted from each layer \( l_i \) in \( L \), with each feature \( f_{l_i} \) having the dimension $\mathbb{R}^{C \times (H/14 \times W/14)}$ for both $I^q \in \mathbb{R}^{3 \times H \times W}$ and $I^s \in \mathbb{R}^{3 \times H \times W}$ because the input image size is divided by the patch size, where the patch size is 14 for the DINO ViT-B/14 model, and C represents the hidden dimension size of the features.  Thus, we obtained a set of features \( F = \{ f_{l_1}, f_{l_2}, \ldots, f_{l_{12}} \} \), capturing multi-level representations across all 12 layers. This comprehensive extraction provides a range of visual information from fine-grained details to high-level abstractions.
Following the MSI process~\cite{msi}, support feature~$F^s=\{f^s_{l_1},f^s_{l_2}, \ldots, f^s_{l_{12}} $ \}, target feature~$F^t=\{f^t_{l_1},f^t_{l_2}, \ldots, f^t_{l_{12}} $ \}, and query feature ~$F^q=\{f^q_{l_1},f^q_{l_2}, \ldots, f^q_{l_{12}} $ \} were extracted from the 12 encoder layers respectively. The extraction of $F^t$ utilized a masked support image~($I^m \in  \mathbb{R}^{ 3 \times H \times W} $), isolating and preserving information pertinent solely to the target:
\begin{align}  \label{eq:2}
  I^{m} & =  I^s \odot \psi (M^s), 
\end{align}
Where $\psi$ resizes the support mask~($M^s \in \mathbb{R}^{H \times W}$) to match the dimensions of $I^s\in  \mathbb{R}^{ 3 \times H \times W} $, then we use the Hadamard product~($\odot$) to eliminate the background. $F^s$ were extracted from the entire support image~$I^s$ without applying a support mask~$M^s$. This approach allows us to include all support information, enabling a comparison with $F^q$ to identify target information not available  in $F^t$.


\subsection{Fully Connected Correlation~(FCC)}
Features extracted from the DINOv2-B~\cite{oquab2023dinov2} are used to calculate $FCC \in \mathbb{R}^{n^2 \times H/14 \times W/14 \times H/14 \times W/14}$ through cosine similarity between $f^t_{l_i} \in F^t$ and $f^q_{l_j} \in F^q$ :
\begin{equation}
FCC(i,j)= \cos(f^t_{l_i}, f^q_{l_j}) = \bigg(\frac{ (f^t_{l_i})^T \cdot f^q_{l_j}}{\left\| f^t_{l_i} \right\| \left\| f^q_{l_j} \right\|}\bigg),
\end{equation}
where \( i \) and \( j \) range from 1 to 12, representing features from layers \( l_1 \) to \( l_{12} \). We used ViT-B~\cite{vit} as the backbone, which comprises 12 layers, but this approach can be extended to accommodate \( n \) layers for other models. Fig.~\ref{fig:method}(a) illustrates the functioning of FCC. After this, we have 144 channels in total for across 12 layers. For \( n \) layers, this would result in $n^2$ channels.  Since output of each layer in the ViT~\cite{vit} has consistent channel and spatial dimensions, we can achieve FCC calculation directly. This operation wasn’t feasible with a ResNet~\cite{Resnet} backbone, which requires resizing and adjusting channel sizes.


\noindent\textbf{Motivation:}
In one-shot segmentation, accurately identifying the target object in the query image is challenging, especially when there are variations in shape, texture, or scale between the support and query images (Fig.~\ref{fig:4cases}). Traditional approaches rely on same-layer correlations, assuming that each layer captures a particular pattern relevant to the target. However, as illustrated in Fig.~\ref{fig:cka_comp_scalediff}(a), Vision Transformers~(ViTs) display a uniform pattern in feature similarity across layers, suggesting that the relationships between layers are interconnected rather than hierarchically distinct. Consequently, relying solely on same-layer correlation provides limited information.

Higher layers retain consistent, scale-invariant features despite object variations, resulting in similar patterns across all Centered Kernel Alignment (CKA~\cite{pmlr-v97-kornblith19a,cka2}) similarity heatmaps. In contrast, lower and middle layers capture varying scales and local details, leading to a broader spread in the similarity heatmap across these layers~(Fig.~\ref{fig:cka_comp_scalediff}). This pattern indicates that cross layer correlations can provide complementary information capturing both detailed, scale dependent features and abstract, object invariant representations. By leveraging these cross-layer correlations, Fully Connected Correlation~(FCC) enriches the segmentation model's ability to recognize target objects with varying appearances, leading to improved segmentation accuracy.



\subsection{Dual-Condition FCC~(DCFC)}

We designed the Dual-Condition Fully Connected Correlation~(DCFC $\in \mathbb{R}^{ (2*n^2)/4 \times H/14 \times W/14 \times H/14 \times W/14}$) to extend FCC. Inspired by MSI~\cite{msi}, we designed two pathways to extract features from the support set~(Fig.~\ref{fig:method}). One from the support mask~$I^s$ and the other from the masked support image~$I^m$. This approach provides a strong signal containing target-specific features~$F^t$, helping the network uncover hidden information within the features~$F^s$. Additionally, FCC is equipped for the two paths to capture fine-grained details across all layers.
\begin{equation}
  \begin{aligned}  \label{eq:3}
    FCC_{t}(i,j)=  \cos(f^t_{l_i}, f^q_{l_j}),\\
    FCC_{s}(i,j)=\cos(f^s_{l_i}, f^q_{l_j})
  \end{aligned}
\end{equation}
where $f^t_{l_i}$ and $f^s_{l_i}$ denote the target and support features, respectively, while $f^q_{l_j}$ represents the query features.  $i$ and $j$ indicate the layer positions within the backbone.
\begin{align} \label{eq:4}
 DCFC =  Conv(FCC_{t} \oplus FCC_{s}),
\end{align}
where $\oplus$ denotes the concatenation. We obtain $FCC_t$ and $FCC_s$ using Equation 3, concatenate them, and then pass the result through a 1x1 convolution layer to reduce the channel size. Conv1x1 convolution reduces computational overhead while allowing the network to learn meaningful representations by capturing relationships across channels.



\subsection{Decoder}

We adopted a lightweight decoder built with a depth-wise separable 4D convolution module~(DSCM)~\cite{dscm}, as it effectively handles large embeddings, which is essential for our FCC. DCFC, the reduced correlation maps by 1x1 Conv, is fed into the decoder which has been adjusted to accommodate the modified channel dimensions.




\section{Experiments}

\noindent\textbf{Datasets:}
To evaluate the proposed method, FCC, we employed the PASCAL-$5^i$~\cite{pascal,pascal_2} and COCO-$20^i$~\cite{lin2015microsoft} datasets. We applied 4-fold testing to validate the approach, ensuring consistency with prior studies~\cite{HSNet,vrpsam,msi,hmmasking}. The classes are divided into four separate groups for training and testing purposes. For each fold, 5 classes are used for testing on PASCAL-$5^i$ and 20 classes on COCO-$20^i$. The remaining classes are used for training. This arrangement guarantees that none of the training classes are included in the test set. For the generalizability test, we train the model on COCO-$20^i$ and evaluate it on PASCAL-$5^i$. To prevent class overlap, we adjust the order of PASCAL classes in line with prior studies~\cite{HSNet,hmmasking,msi}.

\noindent\textbf{Implementation Details:}
We trained FCC using two H100 GPUs. During training, only the FCC module is fine-tuned, with pretrained image backbones frozen to isolate their effect. For the PASCAL-$5^i$~\cite{pascal}, we set the learning rate to 0.001, trained for 100 epochs, and used a batch size of 20. For the COCO-$20^i$~\cite{lin2015microsoft}, we trained for 50 epochs while keeping the other settings the same as for PASCAL-$5^i$. We used the AdamW optimizer~\cite{Adam} and resized all images to 420x420 pixels. For the loss function, we combined dice loss with cross-entropy loss. At each epoch, we evaluated the test accuracy to identify the best-performing model. For the baseline model, we exclude the cross-layer correlation and the DCFC component, utilizing only same-layer correlation.

\noindent \textbf{K-Shots$>$1:} We obtain \textit{K} corresponding mask predictions from the models with a query image and \textit{K} support-set images. All predictions are summed and normalized by the highest score~\cite{HSNet,ASNet,VAT2}.

\noindent\textbf{Evaluation Metrics:}
We report OSS performance using mean Intersection-over-Union~(mIoU) which is widely used in OSS research~\cite{FSOT,PFENet,ASGNet,HSNet,ASNet,VAT2,hmmasking}. We calculate mIoU $=  \frac{1}{n}\sum_{1}^{n}IoU$ where $n$ is the number of test cases.

\begin{table}
\centering
\scalebox{0.45}{
\resizebox{\textwidth}{!}{%
\begin{tabular}{@{}cc|ccccc@{}}
\toprule
\multirow{2}{*}{Backbone} & \multirow{2}{*}{Methods} & \multicolumn{5}{c}{1-shot} \\
 &  & $5^0$ & $5^1$ & $5^2$ & $5^3$ & mIoU \\\midrule
\multirow{8}{*}{ResNet50~\cite{Resnet}} 

 & HSNet~\cite{HSNet} & 64.3 & 70.7 & 60.3 & 60.5 & 64.0  \\

& DCAMA~\cite{DCAMA} & {67.5} & {72.3} & {59.6} & 59.0 & 64.6  \\

 & ASNet~\cite{ASNet} & {68.9} & {71.7} & {61.1} & {62.7} & {66.1}   \\
 
 & VAT~\cite{VAT} & 67.6 & 71.2 & 62.3 & 60.1 & {65.3} \\

& MSI~\cite{msi} & {71.0} & {72.5} & {63.8} & {65.9} & {68.3}   \\

& VRP-SAM~\cite{vrpsam} & 73.9 & 78.3 & 70.6 & 65.0 &  71.9 \\

& LLaFS\textsuperscript{\textdagger}~\cite{llafs} & 74.2 & 78.8 & 72.3 & 68.5 &  73.5 \\

& PI-CLIP\textsuperscript{\textdagger}~\cite{piclip} & 76.4 & 83.5 & 74.7 & 72.8 &  76.8 \\

\midrule

 \multirow{3}{*}{ViT-B~\cite{vit}} 
& FPTrans~\cite{FPTrans} & 67.1 & 69.8 & \textbf{65.6} & 56.4 & 64.7 \\
& DSCM~\cite{dscm} & 67.4 & 73.1 & 63.8 & 63.7 & 67.0 \\

 & \cellcolor{gray!20} \textbf{FCC} & \cellcolor{gray!20} \textbf{71.6} & \cellcolor{gray!20} \textbf{75.8} & \cellcolor{gray!20} 65.0 & \cellcolor{gray!20} \textbf{70.3} & \cellcolor{gray!20}  \textbf{70.7}  \\
 
\midrule

 \multirow{4}{*}{DeiT-B~\cite{deit}} 
 & FPTrans~\cite{FPTrans} & \textbf{72.3} & 70.6 & {68.3} & 64.1 & 68.8 \\
  & MuHS~\cite{MuHS} & {71.2} & 71.4 & {67.0} & 66.6 & 69.1 \\

& DSCM~\cite{dscm} & 67.0 & 74.8 & 64.8 & 64.5 & 67.8 \\

 & \cellcolor{gray!20} \textbf{FCC} & \cellcolor{gray!20} 70.2 & \cellcolor{gray!20} \textbf{77.1} & \cellcolor{gray!20} \textbf{68.7} & \cellcolor{gray!20} \textbf{70.7} & \cellcolor{gray!20} \textbf{71.7}  \\

 \midrule
\multirow{3}{*}{DINOv2~\cite{oquab2023dinov2}} 

& DSCM~\cite{dscm} & 76.5 & 81.3 & 72.1 & 77.4 & 76.8  \\

& DSCM-DFN\textsuperscript{\textdagger}~\cite{dscm} & 78.1 & 83.2 & 76.9 & 80.6 & 79.7  \\

 &  \cellcolor{gray!20} \textbf{FCC} & \cellcolor{gray!20} \textbf{80.5} & \cellcolor{gray!20} \textbf{84.1} & \cellcolor{gray!20} \textbf{77.6} & \cellcolor{gray!20} \textbf{81.7}  & \cellcolor{gray!20} \textbf{81.0}  \\

\bottomrule
  
\end{tabular}
}
}
\caption{Performance evaluation on PASCAL-5$^i$~\cite{pascal}. The best results are shown in \textbf{bold}. \textsuperscript{\textdagger} indicates methods that utilize class awareness for vision language encoder.} 
\label{table:performance_pascal}

\end{table}
    

\begin{table}
  \centering
\scalebox{0.45}{
\resizebox{\textwidth}{!}{%
\begin{tabular}{@{}cc|ccccc@{}}
\toprule
\multirow{2}{*}{Backbone} & \multirow{2}{*}{Methods} & \multicolumn{5}{c}{1-shot}  \\
 &  & $20^0$ & $20^1$ & $20^2$ & $20^3$ & mIoU  \\ \midrule

\multirow{8}{*}{ResNet50~\cite{Resnet}} 
& HSNet~\cite{HSNet} & 36.3 & {43.1} & 38.7 & 38.7 & 39.2   \\

& VAT~\cite{VAT} & {39.0} & 43.8 & 42.6 & {39.7} & {41.3}  \\

& ASNet~\cite{ASNet} & {41.5} & 44.1 & 42.8 & {40.6} & {42.2}   \\   

& BAM~\cite{BAM} & {43.4} & {50.6} & {47.5} & {43.4} & {46.2}   \\

& MSI~\cite{msi} & {42.4} & {49.2} & {49.4} & {46.1} & {46.8}   \\

& VRP-SAM~\cite{vrpsam} & 48.1 & 55.8 & 60.0 & 51.6 &  53.9  \\

& LLaFS\textsuperscript{\textdagger}~\cite{llafs}  & 47.5 & 58.8 & 56.2 & 53.0 & 53.9   \\

& PI-CLIP\textsuperscript{\textdagger}~\cite{piclip}  & 49.3 & 65.7 & 55.8 & 56.3 & 56.8  \\

\midrule  

\multirow{5}{*}{DINOv2~\cite{oquab2023dinov2}} 
& Matcher~\cite{liu2023matcher} & 52.7 & 53.5 & 52.6 & 52.1 & 52.7   \\

& DSCM~\cite{dscm} & 56.0 & 61.3 & 57.9 & 58.8 & 58.5   \\

& VRP-SAM~\cite{vrpsam} & 56.8 & 61.0 & 64.2 & 59.7 &  60.4   \\

& DSCM-DFN\textsuperscript{\textdagger}~\cite{dscm} & 59.1 & 64.5 & 62.5 & 62.7 & 62.2   \\

& \cellcolor{gray!20} \textbf{FCC} 
  & \cellcolor{gray!20} \textbf{62.7} & \cellcolor{gray!20} \textbf{67.4} & \cellcolor{gray!20} \textbf{66.2} 
    & \cellcolor{gray!20} \textbf{66.4} & 
    \cellcolor{gray!20} \textbf{65.7}   \\

\bottomrule
\end{tabular}
}
}
\caption{Performance evaluation on COCO-20$^i$~\cite{lin2015microsoft}. The best results are shown in \textbf{bold}. \textsuperscript{\textdagger} indicates methods that utilize class awareness for vision-language encoder.}
\label{table:performance_coco}
\end{table}

\begin{table}
\centering
\scalebox{0.75}{%
\begin{tabular}{@{}cc|c|c@{}}
\toprule
\multirow{2}{*}{Backbone} & \multirow{2}{*}{Methods} & \multicolumn{1}{c|}{1-shot} & \multicolumn{1}{c}{5-shot}  \\
&  &  mIoU &  mIoU    \\ \midrule

\multirow{1}{*}{ResNet50~\cite{Resnet}} 
 & VRP-SAM~\cite{vrpsam} & {75.9} & {-}\\
 \midrule
\multirow{3}{*}{ResNet101~\cite{Resnet}} 
 & HSNet~\cite{HSNet} & 61.6  & {68.7} \\
 
   & VAT~\cite{VAT} & 64.5 & 69.7 \\  

  & MSI~\cite{msi}&  {67.8} & {72.6} \\  
  
   \midrule
  
\multirow{2}{*}{DINOv2~\cite{Resnet}} 
 & DSCM~\cite{dscm} & {78.4} & {82.4}\\
& \cellcolor{gray!20} \textbf{FCC} & \cellcolor{gray!20} \textbf{81.7} & \cellcolor{gray!20} \textbf{84.6}\\

\bottomrule
\end{tabular}
}
\caption{Generalizability evaluation on PASCAL-5$^i$~\cite{pascal} after training on COCO-20$^i$~\cite{lin2015microsoft}. Notably, the classes used for training on COCO-$20^i$ do not overlap with the classes present in the PASCAL-$5^i$ test set for this study.}
\label{table:performance_pascal_shift}
\end{table}

\begin{table}[!ht]
\centering
\scalebox{0.8}{
\begin{tabular}{l|c|cc}
\toprule
\multirow{2}{*}{Method} & \multirow{2}{*}{Venue} & \multicolumn{2}{c}{{PASCAL-5\textsuperscript{i}}} \\
 & & {1-shot} & {5-shot} \\
 \midrule
\textit{generalist model} \\
Painter*~\cite{Painter} & CVPR'23 & 64.5 & 64.6 \\
SegGPT*~\cite{seggpt} & ICCV'23 & 83.2 & \textbf{89.8} \\
\midrule
\textit{specialist model} \\
HSNet \cite{HSNet} & ICCV'21 & 66.2 & 70.4 \\
HSNet* & & 68.7 & 73.8 \\
VAT \cite{VAT} & ECCV'22 & 67.9 & 72.0 \\
VAT* & & 72.4 & 76.3 \\


FCC  & this work & 81.0 & 84.2 \\
FCC* &  & \textbf{86.8} & 88.2 \\

\bottomrule
\end{tabular}
}
\caption{Performance comparison between specialist and generalist models on PASCAL-5\textsuperscript{i}~\cite{pascal} * indicates that the categories in training cover the categories in testing. Generalist models trained on multiple datasets, not limited to PASCAL-5$^i$.}

\label{table:generalist}

\end{table}

\subsection{Results}

\begin{figure}[!htp]
  \centering
    \includegraphics[height=6.3cm]{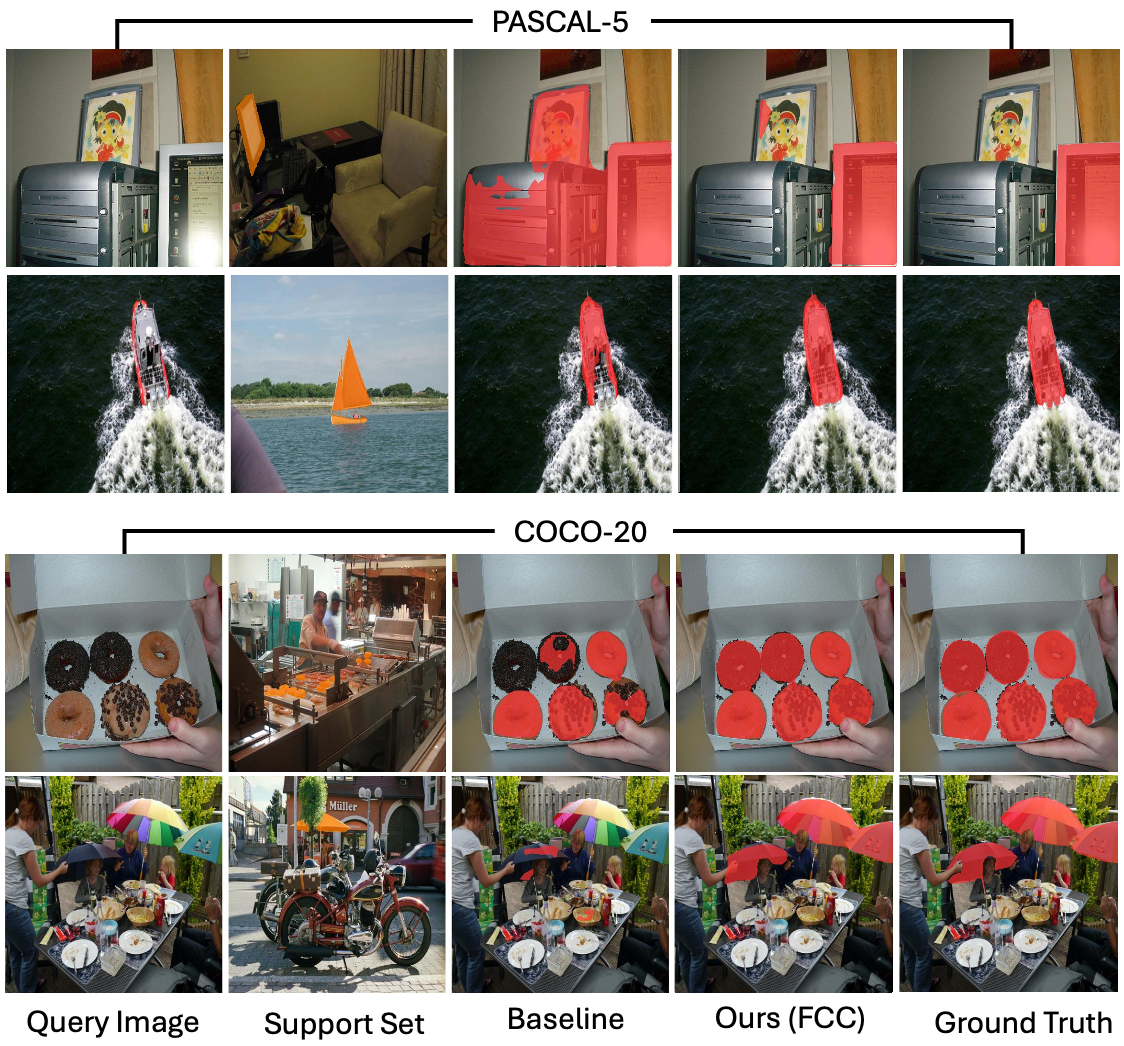}
    \caption{The most challenging cases, where the support mask occupies less than 5\% of the image in PASCAL-5$^i$~\cite{pascal} and COCO-20$^i$~\cite{lin2015microsoft}. Ours~(FCC) with DINOv2~\cite{oquab2023dinov2} shows insightful performance compared to Baseline in one-shot setting.
    }    
    \label{fig:visual_comparison}
\end{figure}

\noindent\textbf{Qualitative Results:} 
Fig.~\ref{fig:visual_comparison} presents a qualitative comparison between the Baseline~\cite{dscm} and FCC. FCC demonstrates clear improvements, particularly in the areas where the baseline struggles to segment the target object. We observed that FCC can achieve accurate segment consistently even in the cases where the support mask size is significantly larger or smaller than the query mask. Additional qualitative results can be found in Supplementary Material Section 1.2.

\noindent\textbf{PASCAL-$5^i$}~\cite{pascal}\textbf{:}
Table~\ref{table:performance_pascal} presents the performance of FCC and SOTA on the PASCAL-5$^i$. FCC demonstrates a substantial improvement over all other models, both vision-only and vision-language models. We test various transformer backbones such as DeiT-B/16 and ViT-B/16. The results consistently demonstrate improvements in performance across all backbones. FCC outperforms other transformer-based models FPTrans~\cite{FPTrans}, MuHS~\cite{MuHS}, and DSCM~\cite{dscm} for DeiT-B/16~\cite{deit} and ViT-B/16~\cite{vit} backbones.

\noindent\textbf{COCO-$20^i$}~\cite{lin2015microsoft}\textbf{:}
Table~\ref{table:performance_coco} illustrates the performance of FCC on the COCO-$20^i$. FCC demonstrates significant improvements compared to previous models.

\noindent\textbf{Generalization Test:}
We evaluated our model's generalization by training it on the COCO-$20^i$ and testing it on the PASCAL-$5^i$~(Table~\ref{table:performance_pascal_shift}). The results show that the high performance achieved on COCO-$20^i$ effectively transfers to PASCAL-$5^i$. Notably, the classes used for training on COCO-$20^i$ do not overlap with the classes present in the PASCAL-$5^i$ test set for this study.

\noindent\textbf{Comparison with Generalist Methods:}
Table~\ref{table:generalist} shows the performance comparison between FCC, generalist models, and other specialized models on the PASCAL-$5^i$. In this study, FCC was retrained on the training data, including test classes. The results showed that FCC outperforms all state-of-the-art specialized methods and generalist models in 1-shot setting. In 5-shot setting, FCC is the second-best model followed by SegGPT (88.2 mIoU vs 89.8 mIoU). However, it is important to know that SegGPT was trained on multiple datasets, not limited to PASCAL-$5^i$.

    


\subsection{Ablation Study and Analysis}

\noindent\textbf{Why FCC is useful:}
FCC benefits from a mix of high and low feature similarity: High similarity in higher layers ensures the capture of semantic consistency and abstract features. Low-to-mid similarity highlights scale-dependent and detailed local features, allowing the model to adapt to variations.
By integrating cross-layer correlations, FCC leverages this balance, combining detailed, scale-sensitive information from lower layers and abstract, invariant information from higher layers. This synergy results in comprehensive segmentation performance, as demonstrated in challenging scenarios like occlusions and shape differences~(Fig.~\ref{fig:4cases}). Further correlation analysis is available in Supplementary Material Section 1.1.


\noindent\textbf{(1)~Visualization:} We used Centered Kernel Alignment~(CKA~\cite{pmlr-v97-kornblith19a,cka2}) similarity heatmaps~(Fig.~\ref{fig:cka_comp_scalediff}) to analyze feature similarity across layers for objects at different shapes. The results suggest that higher layers capture stable scale invariant information while lower layers focus on finer scale sensitive details. This pattern occurs because low to middle layers are tuned to capture localized textures, edges, and patterns that change with object differences, resulting in more dispersed similarity across these layers. In contrast, higher layers capture more abstract and semantic meaning, even under scale variation. These insights confirm FCC’s essential role in leveraging both detailed and abstract information through cross-layer correlations, effectively enhancing segmentation performance when an object's shape varies between support and query images.

\noindent\textbf{(2)~Ablation Study:} We investigated the impact of FCC and DCFC individually. Table~\ref{abstudy_fcc} reveals that incorporating FCC consistently enhances segmentation quality. When both FCC and DCFC were combined, the mIoU reached a peak of 81.0, indicating that the baseline model~\cite{dscm} does not fully leverage the information FCC provides.

\begin{table}
\centering
\scalebox{0.8}{
\begin{tabular}{c|ccc|c}
\hline
\makecell{Method \& \\ Backbone} & {Baseline} & {FCC} & {DCFC}  & {mIoU} \\ \hline
\multirow{4}{*}{\makecell{FCC \\ (DINOv2)}} 
& \checkmark & $\cdot$  & $\cdot$ & 76.8 \\ \cline{2-5} 
& \checkmark & \checkmark  & $\cdot$ & 79.2 \\ \cline{2-5} 
& \checkmark & $\cdot$  & \checkmark & 80.2 \\ \cline{2-5} 
& \checkmark & \checkmark  & \checkmark & \textbf{81.0} \\ \cline{2-5} \hline
\end{tabular}
}
\caption{Ablation study about FCC and DCFC on PASCAL-5$^i$~\cite{pascal}}
\label{abstudy_fcc} 
\end{table}



\begin{figure}[!htp]
  \centering
    \includegraphics[height=4.1cm]{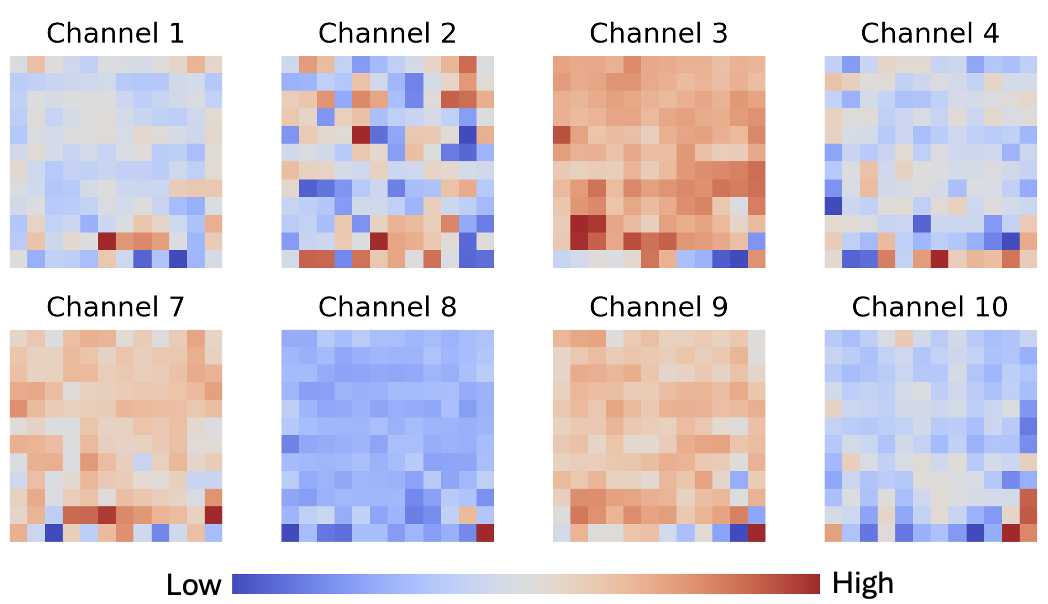}
    \caption{Heatmap visualization of correlation prioritization through 1x1 convolution weights in FCC. The heatmap shows how specific correlations are emphasized in different channels. This selective weighting directs FCC’s focus to the most relevant features, enhancing segmentation accuracy in complex scenarios.}
    
    \label{fig:channel_weight}
\end{figure}

\noindent\textbf{(3)~Correlation Prioritization through 1x1 Conv:} Fig.~\ref{fig:channel_weight} illustrates correlation prioritization through the 1x1 convolution weights, highlighting how FCC effectively utilizes diverse correlations to segment the target object. This visualization demonstrates the usefulness of FCC, leveraging both same-level with cross-level correlations to obtain comprehensive target information. The full heatmap is provided in Supplementary Fig.~7.




\noindent\textbf{The Computation Cost of FCC:} 
We present the FLOPs, memory usage, and inference/training time in Table~\ref{tab:runtime_comparison}, measured on a single H100 GPU with batch sizes of 1 and 4. Although FCC requires increased computational overhead over the baseline, it delivers substantial performance gains, achieving 40 FPS, maintaining a smaller parameter than VRP-SAM~\cite{vrpsam}. VRP-SAM is a cutting-edge model recently introduced. We did not employ parallel computing to accelerate correlation calculations, leaving room for further optimization of cross-correlation computations through parallelization.

\noindent\textbf{Determining the Optimal Number of Layers:} To determine the optimal number of layers for calculating correlations, we performed extensive experiments using various layer combinations~(Table~\ref{table:cross_level_ab}). The numbers indicate the range of layers used for cross-layer correlation calculation. For instance, Cross3 includes the comparison between the same layer, one above, and one below the current level of layer. Ultimately, we found that fully utilizing cross-correlation provides the best performance.

\noindent\textbf{The Most Challenging Cases:}
Table~\ref{table:Ablation_study_small} demonstrates that FCC significantly outperforms the baseline, especially in the most challenging scenarios. Notably, it improves performance on small size support masks by 7.3 mIoU on PASCAL-$5^i$ and 9.9 on COCO-$20^i$, which is considerably higher than the overall improvements of 3.6 mIoU on PASCAL-$5^i$ and 7.2 mIoU on COCO-$20^i$. This indicates that FCC enhances segmentation quality overall, with substantial gains in difficult cases.

\begin{table}[!ht]
\centering 
\resizebox{0.47\textwidth}{!}{ 
\begin{tabular}{@{}c|c|c|c|c|c@{}} 
\toprule 
\multirow{2}{*}{\begin{tabular}[c]{@{}c@{}} Method \\ \end{tabular}} &  
\multirow{2}{*}{\begin{tabular}[c]{@{}c@{}} Params \\ (M) \end{tabular}} &  
\multirow{2}{*}{\begin{tabular}[c]{@{}c@{}}FLOPs \\ (GMac) \\  \end{tabular}} &   
Memory &  
\multirow{2}{*}{\begin{tabular}[c]{@{}c@{}}Inference \\ (Sec~/~Image) \end{tabular}} &  
Training \\  
& & & (GB) & & (Sec~/~Batch) \\ 
\midrule  
VRP-SAM~[\textcolor{RoyalBlue}{38}] & 642.7 & 255.8 & 4.9 & 0.015 & 0.038 \\    
\midrule  
Baseline  & 87.2 & 169.1 & 2.5 & 0.011 &  0.013 \\  
FCC & 87.3 & 246.4 & 4.0 & 0.025 & 0.057 \\  
\midrule  
\end{tabular}} 
\caption{Computational cost on PASCAL-5$^i$ with DINOv2} 
\label{tab:runtime_comparison} 
\end{table}
\vspace{-1.5em}

\begin{table}[!ht]
  \centering
    \begin{minipage}{.47\textwidth}
\centering
\resizebox{\textwidth}{!}{
\begin{tabular}{@{}cc|cccc|c}
\toprule
\multirow{2}{*}{Backbone} & \multirow{2}{*}{Methods} & \multicolumn{5}{c}{1-shot} 
\\
 &  & $5^0$ & $5^1$ & $5^2$ & $5^3$ & mIoU   \\ \midrule
\multirow{7}{*}{DINOv2~\cite{Resnet}} 
&Baseline & 76.5 & 81.3 & 72.1 & 77.4 & 76.8  \\ 
&FCC~(Cross3) & 79.2 & \underline{82.8} & 74.2 & 77.7 & 78.5 \\ 
&FCC~(Dilated Cross3) & 79.4 & 82.7 & 73.1 & {78.4} & {78.4} \\ 
&FCC~(Cross3*) & 77.8 & 80.6 & 72.7 & 76.3 & 76.9 \\ 
&FCC~(3~Nearby 2~Faraway) & \underline{79.7} & 82.3 & \underline{74.3} & \textbf{78.9} & \underline{78.8} \\ 
&FCC~(Cross5) & 79.6 & 82.7 & \underline{74.3} & 77.7 & 78.6  \\ 
&FCC~(Fully Cross) & \textbf{80.4} & \textbf{83.0} & \textbf{74.7} & \underline{78.5} & \textbf{79.2} \\





\bottomrule 
\end{tabular}

}
    \end{minipage}
    \caption{\small Training FCC on PASCAL-5$i$~\cite{pascal}. For this experiment, we removed DCFC. The number specifies the number of layers included in the cross-layer correlation calculation. * indicates that same-layer correlation is excluded.}

\label{table:cross_level_ab}

  \end{table}

\begin{table}[!ht]
  \centering
    \begin{minipage}{.38\textwidth}
\centering
\resizebox{\textwidth}{!}{
\begin{tabular}{@{}cc|cccc|c}
\toprule
\multirow{2}{*}{Backbone} & \multirow{2}{*}{Methods} & \multicolumn{5}{c}{1-shot} 
\\
 &  & $F^0$ & $F^1$ & $F^2$ & $F^3$ & mIoU   \\ \midrule
\multirow{2}{*}{\makecell{DINOv2~\cite{oquab2023dinov2} \\ (PASCAL-5$^i$)}} 
 & Baseline & 73.8 & 63.9 & 61.5 & 61.0 & 65.1 \\
& FCC& \textbf{78.6} & \textbf{69.2} & \textbf{68.7} & \textbf{72.9} & \textbf{72.4}  \\
 
 \midrule 
\multirow{2}{*}{\makecell{DINOv2~\cite{oquab2023dinov2} \\ (COCO-20$^i$)}} 
& Baseline & 51.6 & 50.0 & 55.2 & 49.1 & 51.5 \\
 & FCC & \textbf{60.1} & \textbf{61.2} & \textbf{63.5} & \textbf{60.6} & \textbf{61.4} \\

\bottomrule 
\end{tabular}

}

    \end{minipage}
    \caption{\small Performance comparison on PASCAL-5$^i$~\cite{pascal} and COCO-20$^i$~\cite{lin2015microsoft} for the most challenging cases where support masks occupy below 5\% of the support images. Best results in \textbf{bold}. }
\label{table:Ablation_study_small} 
  \end{table}

\section{Conclusion}

We introduced FCC, a fully connected correlation approach for one-shot segmentation, which delivered an impressive performance, achieving high-level segmentation accuracy. We validated the effectiveness of FCC on the PASCAL-$5^i$ and COCO-$20^i$ datasets, even under domain shift conditions. FCC shows potential for application in other domain-specific tasks that require similar capabilities.

\section{Acknowledgment}
This research used resources of the National Energy Research Scientific Computing Center, a DOE Office of Science User Facility supported by the Office of Science of the U.S. Department of Energy under Contract No. DE-AC02-05CH11231 using the NERSC award NERSC DDR-ERCAP0033558.

{
    \small
    \bibliographystyle{ieeenat_fullname}
    \bibliography{main}
}

\end{document}